\documentclass[10pt,twocolumn,letterpaper]{article}

\usepackage{iccv}
\usepackage{times}
\usepackage{epsfig}
\usepackage{graphicx}
\usepackage{amsmath}
\usepackage{amssymb}
\usepackage{bm}

\usepackage[ruled]{algorithm2e} 
\usepackage{booktabs}
\usepackage{tabularx}
\newcommand{\norm}[1]{\left\lVert#1\right\rVert}

\usepackage[breaklinks=true,bookmarks=false]{hyperref}

\iccvfinalcopy 

\def\iccvPaperID{****} 

\ificcvfinal\fi

\begin{document}

\title{Single-Shot Motion Completion with Transformer}

\author{Yinglin Duan $^*$, Tianyang Shi $^*$\\
NetEase Fuxi AI Lab\\
{\tt\small \{duanyinglin, shitianyang\}@corp.netease.com}
\and
Zhengxia Zou \thanks{Contributed Equally. (Names are in alpha order)}\\
University of Michigan, Ann Arbor\\
{\tt\small zzhengxi@umich.edu}
\and
Yenan Lin, Zhehui Qian, Bohan Zhang\\
NetEase \\
{\tt\small yenan\_lin@foxmail.com,}\\
{\tt\small \{qianzhehui, hzzhangbohan\}@corp.netease.com}
\and
Yi Yuan \thanks{Corresponding author.}\\
NetEase Fuxi AI Lab\\
{\tt\small yuanyi@corp.netease.com}
}

\maketitle

\ificcvfinal\thispagestyle{empty}\fi

\begin{abstract}
Motion completion is a challenging and long-discussed problem, which is of great significance in film and game applications. For different motion completion scenarios (in-betweening, in-filling, and blending), most previous methods deal with the completion problems with case-by-case designs. In this work, we propose a simple but effective method to solve multiple motion completion problems under a unified framework and achieves a new state of the art accuracy under multiple evaluation settings. Inspired by the recent great success of attention-based models, we consider the completion as a sequence to sequence prediction problem. Our method consists of two modules - a standard transformer encoder with self-attention that learns long-range dependencies of input motions, and a trainable mixture embedding module that models temporal information and discriminates key-frames. Our method can run in a non-autoregressive manner and predict multiple missing frames within a single forward propagation in real time. We finally show the effectiveness of our method in music-dance applications. Our animated results can be found on our project page \url{https://github.com/FuxiCV/SSMCT}.
\end{abstract}


\begin{figure}[ht]
 \centering
  \includegraphics[width=\linewidth]{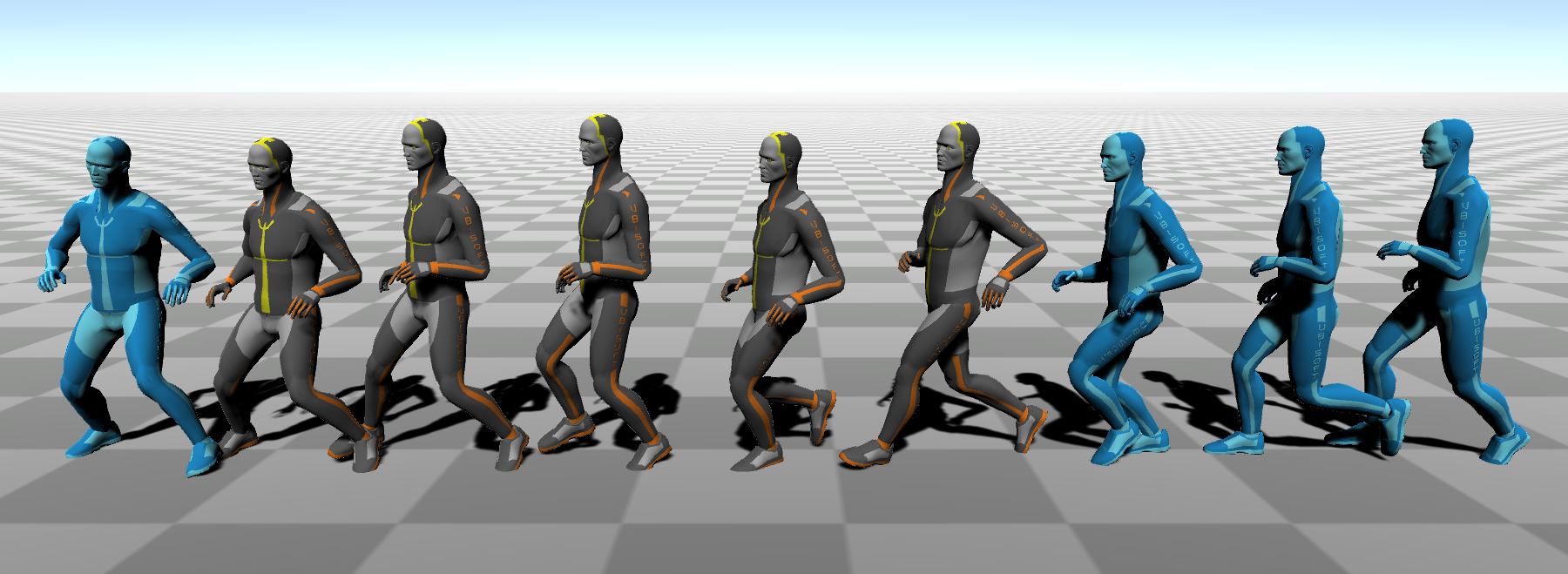}
  \caption{Motion completion by our method based on input keyframes (blue ones). Our method is a unified framework that can solve multiple motion completion problems. We achieve a new state of the art on a high-quality motion completion dataset -- LaFAN1~\cite{harvey2020robust}. 
  }
  \label{fig:teaser}
\end{figure}
\section{Introduction}

Motion completion is an important and challenging problem that has been studied for a long time. Motion completion provides fundamental technical support for animation authoring of 3D characters, and has been recently successfully applied in film production and video games~\cite{thomas2009integrated, ciccone2019tangent}.

In recent years, deep learning methods have greatly promoted the research progress of motion completion.  With recent advances in this field, manpower can now be greatly saved, where high-quality motion can be smoothly generated from a set of historical or sparsely key-frames by learning over a large scale of motion capture data~\cite{kaufmann2020infilling,harvey2020robust}. However, most previous methods deal with the completion problem for different completion scenarios (in-betweening, in-filling, and blending) with case-by-case designs. In this paper, we propose a novel framework that can unify the above processing scenarios. In our method, we leverage the recent popular deep learning architecture named Transformer \cite{vaswani2017attention}, which is built on the self-attention mechanism and now has been widely used in neural language processing, computer vision, and reinforcement learning \cite{tay2020efficient, dosovitskiy2020VIT, parisotto2020stabilizing}. We adopt BERT~\cite{devlin2018bert}, a recent well-known transformer encoder as our backbone, where known frames (key-frames) and unknown frames (need to be complemented) are fed together orderly as input and thus all the frames can be predicted in a single propagation at inference. Our method works in a non-autoregressive manner and can be easily accelerated with GPU parallelization.  As a comparison, most previous methods adopt recursive or auto-regressive prediction where the motion frames need to be generated iteratively that lacks parallelism. 

We consider motion completion in the following three scenarios:

\begin{itemize}
\item[*] \emph{In-betweening.} Animators are required to complement the motion frame-by-frame between the past frames and the provided further keyframe~\cite{harvey2020robust}. 
\item[*] \emph{In-filling.} As an extension of in-betweening, in-filling poses the characters on specific positions of the timeline~\cite{ciccone2019tangent}, and complements the rest of the frames~\cite{kaufmann2020infilling}. A sub-task of In-filling is temporal super-resolution, where the input motion frames are considered as keyframes with equal interval~\cite{harvey2018recurrent}.
\item[*] \emph{Blending.} Blending focuses on the automatic generation of the transition between a pair of pre-defined motions and has been widely used in video games\footnote{\url{https://docs.unity3d.com/Packages/com.unity.timeline@1.6/manual/clp_blend.html}}. For example, many games provide dance fragments for players to choreograph, and blending helps players to concatenate and smooth those chosen fragments.
\end{itemize}

We also introduce a mixture embedding module that further integrates temporal knowledge to the transformer encoder. Our embedding module contains two types of learnable embeddings: position embedding and keyframe embedding. \emph{Position embedding} is a widely studied technology in recent transformer literature, where a set of pre-defined sinusoidal signals are usually used to introduce temporal knowledge to the transformer model~\cite{harvey2018recurrent,harvey2020robust}. In our method, we further make this embedding trainable to deal with different motion completion scenarios. In addition to the encoding of input frame orders, we also introduce \emph{Keyframe embedding} to annotate the input frames and tells the model which parts of the input frames are keyframes (already known) and which frames need to be predicted (unknown). Since the keyframe embedding may have different forms, our method can be easily applied to different completion scenarios regardless of how the input keyframes are arranged. Our design can be also considered as a motion-version of Mask Language Model (MLM)~\cite{taylor1953MLM}, where the unknown frames are represented by a deep bi-directional model trained with self-supervised losses~\cite{devlin2018bert}.

The contributions of our paper are summarized as follows:

1. We investigate the capability of the transformer-based model in the motion completion task. We propose a simple but efficient method to solve the motion completion problems of different application scenarios under a unified framework.

2. Our method works in a parallel prediction manner with high computational efficiency. On a single CPU desktop (I7-8700K @ 3.70GHz), our method can run in real time (40 motion sequences per second, each with 30 frames long).

3. We achieve new in-betweening benchmarking accuracy under multiple evaluation settings. Results on other completion tasks (in-filling and blending) are also reported as baselines for fostering the research community \footnote{We will make our model and pre-trained weights publically available.}.

\section{Related works}
\subsection{Motion completion}

Motion completion is an emerging research hot-spot in computer graphics and multimedia. Motion completion can be viewed as a conditional motion sequence generation problem. Different from those unconditional motion synthesis tasks~\cite{sidenbladh2002implicit,wang2007gaussian,taylor2007modeling} that focus on the generation of unconstrained motion sequences by directly sampling from their posterior distribution, motion completion aims at filling the missing frames in a temporal sequence based on a given set of keyframes. 

Motion completion has a long research history, which can be traced back to the late 1980s. Early works of motion completion typically adopt inverse kinematics to generate realistic transitions between keyframes. For example, space-time constraints and searching based methods were proposed in the 1980s-1990s~\cite{witkin1988spacetime,ngo1993spacetime} to compute optimal physically-realistic trajectory. By using such techniques, transitions between different motions can be smoothly generated~\cite{rose1996efficient}. Also, probabilistic models like the maximum a posterior methods~\cite{chai2007constraint,min2009interactive}, the Gaussian process~\cite{wang2007gaussian}, Markov models~\cite{lehrmann2014efficient} were introduced to motion completion tasks and were commonly used after 2000s. 

Recently, deep learning methods have greatly promoted the research and the performance of motion completion, where the recurrent neural network is the most commonly used framework for motion completion of the deep learning era~\cite{zhang2018data,harvey2018recurrent}. For example, Harvey \etal introduce a novel framework named Recurrent Transition Networks (RTN) to learn a more complex representation of human motions with the help of LSTM~\cite{harvey2018recurrent}. Besides, generative adversarial learning has been also introduced to the motion completion to make the output motion more realistic and naturalistic~\cite{hernandez2019human}. Some non-recurrent motion completion models are also proposed very recently. Kaufmann \etal propose an end-to-end trainable convolutional autoencoder to fill in missing frames~\cite{kaufmann2020infilling}. Another recent progress by Harvey \etal introduces time-to-arrival embeddings and scheduled target-noise to further enhance the performance of RTN and achieve impressive completion results~\cite{harvey2020robust}. Different from the previous RNN-based or convolution-based method, we propose a transformer-based model, which is a more unified solution to this task and can deal with arbitrary missing frames in a single-shot prediction manner. 

\subsection{Motion control} 

Motion control is a typical conditional motion generation task, which is also highly related to motion completion. In motion control, the control signal comes from a pre-defined temporal sequence, e.g. root trajectory, rather than a set of keyframes in motion completion.

Graph-based motion control is the most common type of method in this field before the deep learning era~\cite{lee2002interactive,safonova2007construction,kovar2008motion,arikan2002interactive,beaudoin2008motion}. For example, Arikan \etal formulate the motion generation as a randomized search of the motion graph, which allows to edit complex motions interactively~\cite{arikan2002interactive}. Beaudoin \etal propose a string-based motif-finding algorithm named Motion-Motif Graphs, which further considers the motif length and the number of motions in a motif~\cite{beaudoin2008motion}. Besides, there are many statistical methods proposed to avoid searching from predefined motion templates~\cite{grochow2004style,wang2007gaussian,min2009interactive,min2012motion}. Chai \etal propose a statistical dynamic model to generate motions with motion prior (e.g. user-defined trajectory) and formulate the constrained motion synthesis as a maximum a posterior problem~\cite{chai2007constraint}. Ye \etal introduce a nonlinear probabilistic dynamic model that can handle perturbations~\cite{ye2010synthesis}. Levine \etal propose a probabilistic motion model that learns a low-dimensional space from example motions and generates character animation based on user-specified tasks~\cite{levine2012continuous}.

Recently, deep learning methods have become the mainstream motion control method. Some popular architectures like convolutional autoencoder and recurrent neural network are widely used in this problem~\cite{holden2015learning, holden2016deep, holden2017phase,lee2018interactive}. Adversarial training has also played an important role in motion control and can help generate more realistic motion sequences~\cite{barsoum2018hp, gui2018adversarial, harvey2020robust}. Besides, some recent approaches also leverage reinforcement learning to further incorporate physical rules to improve the quality of the generation results~\cite{coros2009robust, yin2007simbicon,baram2016model, peng2017deeploco,peng2018deepmimic,bergamin2019drecon}.

\begin{figure*}[ht]
  \centering
  \includegraphics[width=0.8\linewidth]{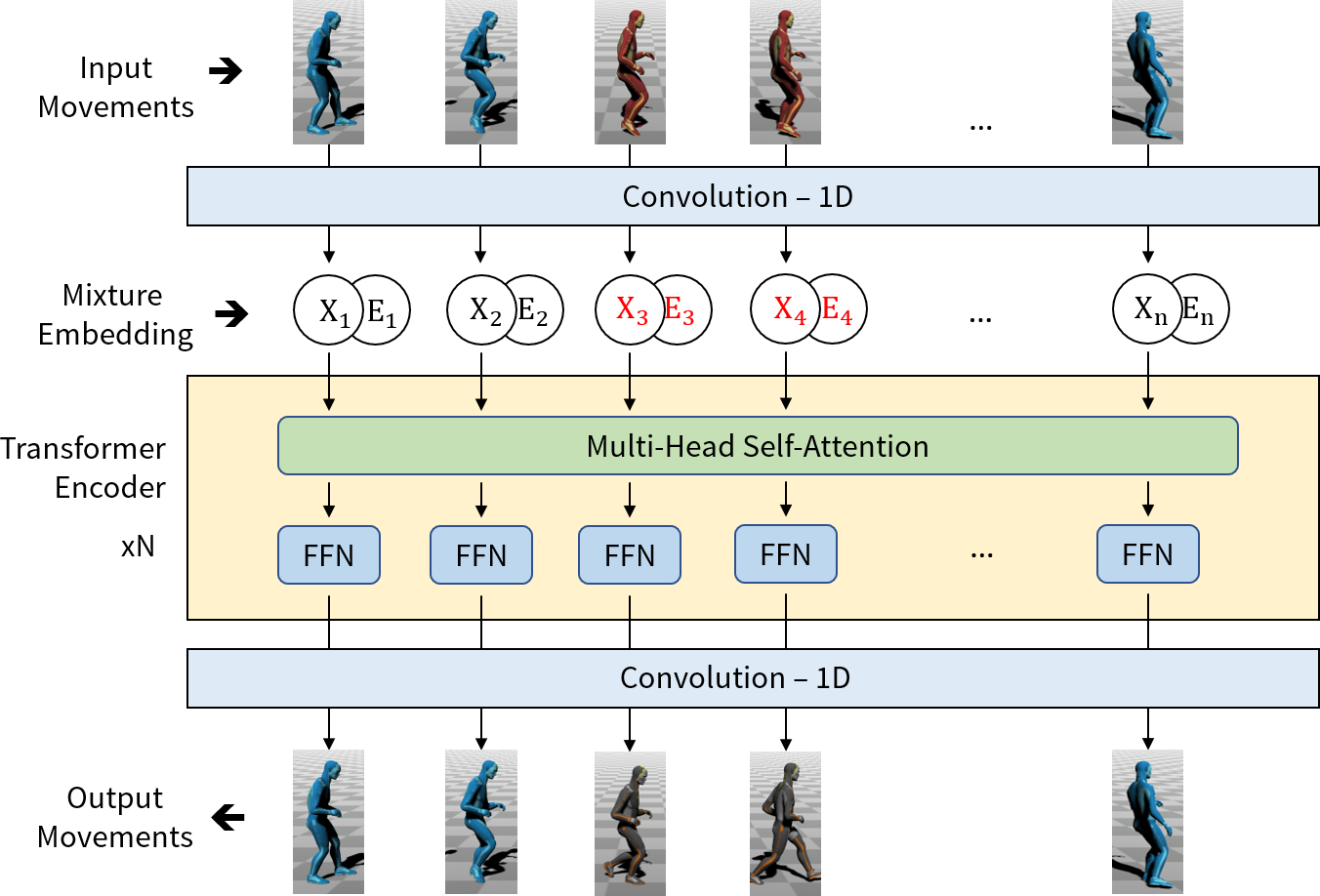}
  \caption{An overview of our method. Our method consists of a standard transformer encoder, a mixture embedding layer and input/output convolutional heads. In motion completion, the unknown input frames are first generated by using linear interpolation ({\color{red} shown in red}) before being fed to the model. Our method takes in a whole masked sequence and complete the prediction within only a single forward propagation.}
  \label{fig:pipeline}
\end{figure*}

\section{Methods}

In this work, we formulate the motion completion as a sequence-to-sequence prediction problem. The unknown motion frames can be generated in a single inference forward propagation conditioned by those input keyframes. We choose BERT, an off-the-shelf transformer architecture~\cite{devlin2018bert} as our network backbone with minimum modifications, and thus the subsequent varieties can be introduced without impediment.

\subsection{Motion completion transformer} \label{sec:mocot}

Fig.\ref{fig:pipeline} shows an overview of our method. Our network consists of 1) a mixture embedding module that converts the motion to a set of sequential tokens, and 2) a standard transformer encoder used to process sequential features. Our method supports multiple input coded format, e.g. [local positions \& rotations] or [global positions \& rotations] or [positions only]. Without loss of generality, we assume that the input has both positions ($x$, $y$, $z$) and rotations ($q_0$, $q_1$, $q_2$, $q_3$) variables (no matter local or global), and therefore, a single pose contains a position coordinate matrix $\bm{P} \in \mathbb{R}^{J\times3}$ and a quaternion matrix $\bm{Q} \in \mathbb{R}^{J\times4}$, where $J$ represents the joint number of the input pose.

For each input pose, we firstly flatten $\bm{P}$ and $\bm{Q}$ into 1-D vectors $\bm{p}$ and $\bm{q}$, and then concatenate the two vectors together into a long vector:
\begin{equation}
\bm{x} = [\bm{p}, \bm{q}] \in \mathbb{R}^{J\times(3+4)}.
\end{equation}
For those unknown input frames, we use the linear interpolation to fill in their missing values along the temporal dimension before feeding their pose vector to the model.

We then use a 1-D temporal convolution layer to transform those pose vectors to a set of ``language tokens'':
\begin{equation}
\bm{Z} = \text{Conv1d}([\bm{x}^1;\bm{x}^2;...;\bm{x}^T])
\end{equation}
where $T$ is the length of the input sequence, $\bm{Z} \in \mathbb{R}^{T\times F}$ is the formated temporal feature of the tokens, and $F$ is the output dimension of the Conv1d layer. The convolution is performed in the joint dimension. Note that different from the previous transformer-based models in computer vision~\cite{dosovitskiy2020VIT} that use a linear projection layer to generate the embeddings, here we use a convolution layer for better capturing the temporal information, e.g. velocity and acceleration.

Considering that the transformer cannot know the order and the exact location of the keyframes in the input sequence, we introduce a mixture embedding $\bm{E}$ to annotate these frames before feeding their features into transformer. For each group of input configuration of keyframes, an embedding $\bm{E}$ is learned as a global variable on the training data and will not change along with the pose feature $\bm{Z}$. We represent the final annotated features $\bm{Z}$ as follows:
\begin{equation}
\bm{\hat{Z}} = [\bm{z}^1 + \bm{e}^1; \bm{z}^2 + \bm{e}^2; ...; \bm{z}^T+ \bm{e}^T]
\end{equation}
where $\bm{z}^t$ and $\bm{e}^t$ are the sub-vectors of input feature $\bm{Z}$ and mixture embedding $\bm{E}$ at the time $t$.

The BERT transformer we used consists of multiple encoder layers~\cite{devlin2018bert}. Each encoder layer further consists of a multi-head self-attention layer (MHSA) and a feed-forward network (FFN). A residual connection~\cite{he2016resnet} is applied across the two layers. The forward mapping of the transformer can be written as follows:
\begin{equation}
\begin{split}
\bm{\hat{H}}^l =&~\text{Norm}(\bm{H}^{l-1} + \text{MHSA}(\bm{H}^{l-1}))\\
\bm{H}^l =&~\text{Norm}(\bm{\hat{H}}^l + \text{FFN}(\bm{\hat{H}}^l))
\end{split}
\end{equation}
where $\bm{H}$ is the output of hidden layers. $l=1,...,L$ are the indices of encoder layers. ``Norm'' represents the layer normalization~\cite{ba2016layer} placed at the output end of the residual connections. We use $\bm{H}^0 = \text{Norm}(\bm{\hat{Z}})$ as the hidden representation of the input layer.

In the multi-head self-attention layer (MHSA), a dense computation between each pair of input frames are conducted, and thus a very long-range temporal relations between frames can be captured. The processing of a single head can be represented as follows: 
\begin{equation}
\bm{f}_{att} = \text{Softmax}(\frac{\bm{Q}\bm{K}^T}{\alpha})\bm{V}
\end{equation}
where $\bm{Q} = \bm{W}_q\bm{H}$ represents a query matrix, $\bm{K} = \bm{W}_k\bm{H}$ represents a key matrix, and $\bm{V} = \bm{W}_v\bm{H}$ represents a value matrix. $\bm{W}_q$, $\bm{W}_k$ and $\bm{W}_v$ are all learnable matrices. We follow the multi-head attention configuration in BERT and set the dimension of $\bm{Q}$, $\bm{K}$ and $\bm{V}$ to $\frac{1}{M}$ of the input $\bm{H}$, $M$ represents the number of heads in the attention layer. Finally, the outputs from different heads are collected, concatenated and projected by a matrix $\bm{W}_{mhsa}$ as the output of the MHSA layer:
\begin{equation}
\text{MHSA}(\bm{H}) = \bm{W}_{mhsa}[\bm{f}_{att}^{(1)}; \bm{f}_{att}^{(2)}; ...; \bm{f}_{att}^{(M)}].
\end{equation}

For the feed-forward network (FFN), it consists of two linear layers and a GeLU layer~\cite{hendrycks2016gelu}. FFN processes each of the frame individually:
\begin{equation}
\text{FFN}(\bm{H}) = \bm{W}_{ffn}^{(1)}(\text{GeLU}(\bm{W}_{ffn}^{(2)}(\bm{H}))),
\end{equation}
where $\bm{W}_{ffn}^{(1)}$ and $\bm{W}_{ffn}^{(2)}$ are the learnable linear projection matrices.
Finally, we apply another 1d-convolution layer at the output end of the transformer, and final completion output $\bm{Y}$ can be written as follows:
\begin{equation}
\bm{Y} = \text{Conv1d}(\bm{H}^N).
\end{equation}

\subsection{Mixture embeddings}

\begin{figure}[t]
  \centering
  \includegraphics[width=1\linewidth]{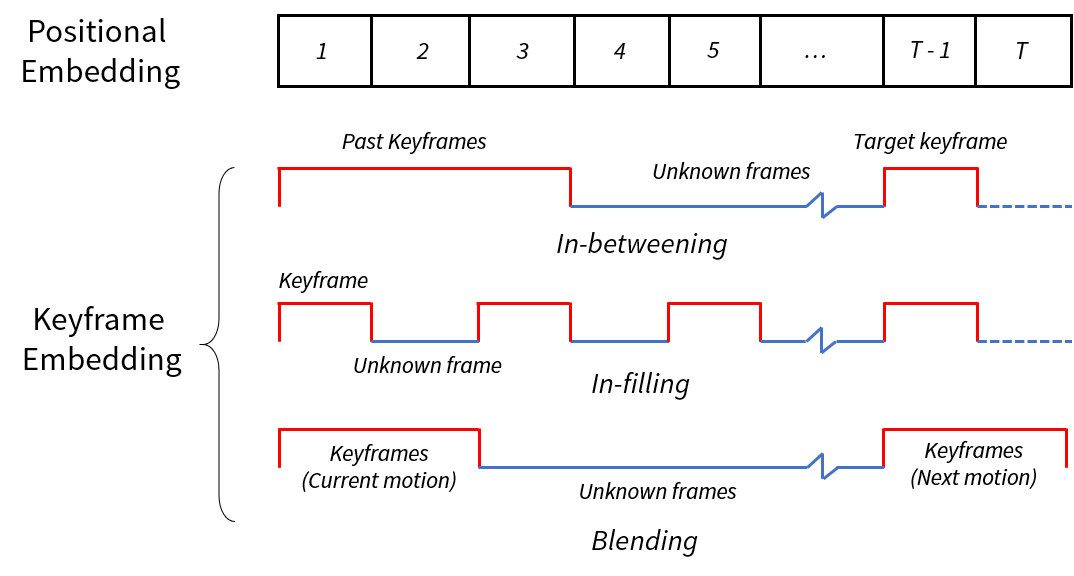}
  \caption{An illustration of mixture embedding module. We design a position embedding and a keyframe embedding, where the former one integrates position information and the latter one annotates whether the frame is the keyframe or not. (Blue dash-lines represent ignored frames that exceeds the prediction range)}
  \label{fig:embeddings}
\end{figure}

The mixture embedding $\bm{E}$ we used consists a positional embedding $\bm{E}_{pos} \in \mathbb{R}^{T\times F}$ and a keyframe embedding $\bm{E}_{kf} \in \mathbb{R}^{T\times F}$, where $T$ is the length of temporal sequence and $F$ is the input feature dimension of transformer. Fig.\ref{fig:embeddings} gives an illustration of the mixture embedding module.

The position embedding $\bm{E}_{pos}$ is a matrix that contains $T$ sub-vectors, each for a single time step: 
\begin{equation}
    \bm{E}_{pos} = [\bm{e}_{pos}^1,\bm{e}_{pos}^2,...,\bm{e}_{pos}^T].
\end{equation}
The keyframe embeddings $\bm{E}_{kf}$ are selected from a learnable  dictionary $\mathcal{D}$ that contains three types of embedding vectors $\mathcal{D} = \{ \hat{\bm{e}}_0, \hat{\bm{e}}_1, \hat{\bm{e}}_2 \}$, which annotate the keyframes, unknown frames, and ignored frames, respectively (Keyframe = 0, Unknown = 1, ignored = 2). These types of frames can be configured in any forms according to different completion tasks and scenarios (e.g., in-betweening, in-filling, and blending). The keyframe embeddings are written as follows:
\begin{equation}
    \bm{E}_{kf} = [\bm{e}_{kf}^1,\bm{e}_{kf}^2,...,\bm{e}_{kf}^T],
\end{equation}
where $\bm{e}_{kf}^{m} \in \{ \hat{\bm{e}}_0, \hat{\bm{e}}_1, \hat{\bm{e}}_2 \}$. Finally, the two types of embeddings are mixed by adding together position-by-position:
\begin{equation}
    \bm{E} = \bm{E}_{pos} + \bm{E}_{kf}.
\end{equation}
It is worth noting that our keyframe embedding is not limited to the above three configurations shown in Fig.\ref{fig:embeddings}. It can be in any format with practical meaning. As a special case, if the keyframes are randomly specified, then our keyframe embedding turns out to be the random token mask in the original BERT paper~\cite{devlin2018bert}.

\subsection{Loss function}

We train our transformer and the embeddings with multi-tasks regression losses. Given a set of predicted motions and their ground truth, we design our pose reconstruction loss $\mathcal{L}_{rec}$ as follows:
\begin{equation}
        \mathcal{L}_{rec} = \frac{1}{NT}\sum_{n=1}^N \sum_{t=1}^T (\norm{\bm{p}_n^t - \hat{\bm{p}}_n^t}_1 + \norm{\bm{q}_n^t - \hat{\bm{q}}_n^t}_1)
\end{equation}
where $\bm{p}^t_n$ and $\bm{q}^t_n$ are the position coordinates and rotation quaternions of the predicted motion sequence $n$ at the time step $t$. $\hat{\bm{p}}^t_n$ and $\hat{\bm{q}}^t_n$ $T$ are their ground truth. $N$ represents the length of motion sequence and the total number of training sequences in the data set. Note that the above losses can be used in both global and local coordinate systems. When applied in the local one, the $\bm{p}_n^t$ can be replaced by $\bm{r}_n^t$, i.e. root coordinates, since the T-pose (or offsets) will always be a constant vector for the same character.

In addition to the above pose reconstruction loss, we also introduce two  kinematics losses to improve the results in different coordinate systems. 1) Forward Kinematics (FK) loss $\mathcal{L}_{FK}$. We follow Harvey \etal~\cite{harvey2020robust} and apply FK loss to our multi-task training loss. The main idea behind the FK loss is to calculate the global position coordinates by using local ones with forward kinematics and weight the local reconstruction loss on different joints:
\begin{equation}
    \mathcal{L}_{FK} = \norm{\text{FK}(\bm{r}, \bm{q}_{local}) - \bm{p}_{global}}_1,
\end{equation}
2). Inverse Kinematics (IK) loss $\mathcal{L}_{IK}$. We also apply IK loss to constrain the T-pose when in global coordinate system. We first compute the local position from the global one with inverse kinematics, and then we compare the offsets between the inverse output and the original input:
\begin{equation}
    \mathcal{L}_{IK} = \norm{\text{IK}(\bm{p}_{global}, \bm{q}_{global}) - \bm{b}}_1
\end{equation}
where $\bm{b}$ represents the offset vector. We remove the root coordinate and keep predicted offsets only when computing the IK loss.

Our final training loss is written as follow:
\begin{equation}
    \mathcal{L} = \alpha_{rec}\mathcal{L}_{rec} + \alpha_{K}\mathcal{L}_{K},
\end{equation}
where $\alpha_{rec}$ and $\alpha_{K}$ are the coefficients to balance different loss terms. $\mathcal{L}_{K}$ represents the kinematics loss in global system ($\mathcal{L}_{IK}$) or in local system ($\mathcal{L}_{FK}$).

\begin{table*}[ht]
\centering
\caption{Experimental results on LaFAN1 dataset. A lower score indicates better performance. (*Note that for a fair comparison, the T-pose of our global results have been replaced by a standard one in local coordinate system.)}
\begin{tabularx}{\textwidth}{l@{\extracolsep{\fill}} ccccccccc}
&\multicolumn{3}{c}{\textbf{L2Q}} &\multicolumn{3}{c}{\textbf{L2P}} &\multicolumn{3}{c}{\textbf{NPSS}} \\
\cmidrule(lr){2-4}\cmidrule(lr){5-7}\cmidrule(lr){8-10}
Length & 5 & 15 & 30 & 5 & 15 & 30 & 5 & 15 & 30\\
\cmidrule(lr){1-10}
Zero-Vel & 0.56 & 1.10 & 1.51 & 1.52 & 3.69 & 6.60 & 0.0053 & 0.0522 & 0.2318 \\
Interp & 0.22 & 0.62 & 0.98 & 0.37 & 1.25 & 2.32 & 0.0023 & 0.0391 & 0.2013 \\
\cmidrule(lr){1-10}
ERD-QV (\cite{harvey2020robust}) & 0.17 & 0.42 & 0.69 & 0.23 & 0.65 & 1.28 & 0.0020 & 0.0258 & 0.1328 \\
Ours (local w/o FK) & 0.18 & 0.47 & 0.74 & 0.27 & 0.82 & 1.46 & 0.0020 & 0.0307 & 0.1487\\
Ours (local)  & 0.17 & 0.44 & 0.71 & 0.23 & 0.74 & 1.37 & 0.0019 & 0.0291 & 0.1430\\
Ours (global w/o ME \& IK) & 0.16 & 0.37 & 0.63 & 0.24 & 0.61 & 1.16 & 0.0018 & 0.0243 & 0.1284\\
Ours (global w/o IK)  & 0.14 & 0.36 & 0.61 & \bf{0.21} & 0.57 & 1.11 & 0.0016 & 0.0238 & 0.1241\\
Ours* (global-full)  & \bf{0.14} & \bf{0.36} & \bf{0.61} & 0.22 & \bf{0.56} & \bf{1.10} & \bf{0.0016} & \bf{0.0234} & \bf{0.1222} \\
\bottomrule
\end{tabularx}%
\label{tab:inbetweening}
\end{table*}%

\subsection{Implementation details}

In our method, we adopt BERT~\cite{devlin2018bert} as the backbone of our transformer with 8 encoder layers. In each encoder layer, we set the number of attention heads to $M=8$. For our input and output Conv1d layers, the kernel size is set to 3 and the padding is set to 1. We set the dimension of the feature embedding in the MHSA layers to 256, and set those in the FFN layers to 512. In our training loss, we set $\alpha_{rec} = 1.0$ and $\alpha_{K} = 0.01$. Consider that the quaternions $\bm{q} \in [0, 1]$ while the position coordinates $\bm{p}$ are in a much larger range, we scale the localization loss and the rotation loss to the same order of magnitude.

We train our network by using Adam optimizer~\cite{kingma2014adam}. We set the maximum learning rate to $10^{-3}$. The whole framework is implemented by using PyTorch~\cite{NEURIPS2019_9015}. For a more detailed training configuration, please refer to our experimental section.  

\begin{table}[ht]
\centering
\caption{Speed performance comparison. CPU inference time are recorded in different batch sizes (1 \& 10) where Inbetweening length is set to 30 frames (i.e. 1 second).}
\begin{tabularx}{1\linewidth}{l@{\extracolsep{\fill}} cc|c}
Method & 1 x 30 & 10 x 30 & CPU info\\
\midrule
ERD-QV~\cite{harvey2020robust} & 0.31s & 0.40s & E5-1650 @ 3.20GHz\\
Ours & \bf{0.025s} & \bf{0.083s} & I7-8700K @ 3.70GHz\\
\bottomrule
\end{tabularx}%
\label{tab:speed}
\end{table}%

\section{Experiments}

\subsection{Datasets and motion completion tasks}

In our experiment, we evaluate our method across three different motion completion tasks:

1. In-betweening on LaFAN1~\cite{harvey2020robust}: LaFAN1 is a public high-quality general motion dataset introduced by Harvey \etal from Ubisoft. In the in-betweening completion task, given the past 10 keyframes and another future keyframe, we aim to predict the motion of the rest frames. 

2. In-filling on Anidance~\cite{tang2018anidance}: Anidance is a public music-dance dataset proposed by Tang \etal~\cite{tang2018anidance}. We test our method on this dataset for the in-filling task, where equally spaced keyframes are given.

3. Blending on our dance dataset: We collect a new dance movement dataset, which contains high-quality dance movements performed by senior dancers and is more challenging than the previous ones. We evaluate this dataset for exploring the potential of our method in dance applications in game environments.

To build our dataset, we follow the classic choreography theory of Doris Humphrey~\cite{humphrey1959art} and define dance phrases as our basic movement unit. We invited four senior dancers to perform five types of dance movements (including Jazz dance, Street dance, J-pop dance, Indian dance, Uygur dance). We use motion capture devices (Vicon V16 cameras) to record the dance movements in 30Hz. Finally, more than 130,000 frames are recorded in our dataset. For convenience, we re-target the dance movements on a standard character released by Harvey \etal~\cite{harvey2020robust}.

\subsection{Metrics}

We follow Harvey \etal~\cite{harvey2020robust} and use L2Q, L2P, and NPSS as our evaluation metrics. The L2Q defines the average L2 distances of the global quaternions between the predicted motions and their ground truth. Similarly, the L2P defines the average L2 distances of the global positions.\footnote{In Harvey \etal's implementation, the predicted positions and their ground truth are normalized by mean and std of the training set. We also follow this setting for a fair comparison.} The NPSS, proposed by Gopalakrishnan~\cite{gopalakrishnan2019neural}, is a variant of L2Q, which computes the Normalized Power Spectrum Similarity and is based on angular frequency distance between the prediction and the groundtruth. 

Note that when we generate our results based on the global coordinate system, we replace the T-pose of these results with the standard one under the local coordinate system (by the simplest IK, i.e. global to local coordinate transformation) for a fair comparison. We found this operation may slightly reduce our accuracy. We will give more discussion on this interesting observation in our Discussion section.

\begin{figure*}[ht]
  \centering
  \includegraphics[width=0.9\linewidth]{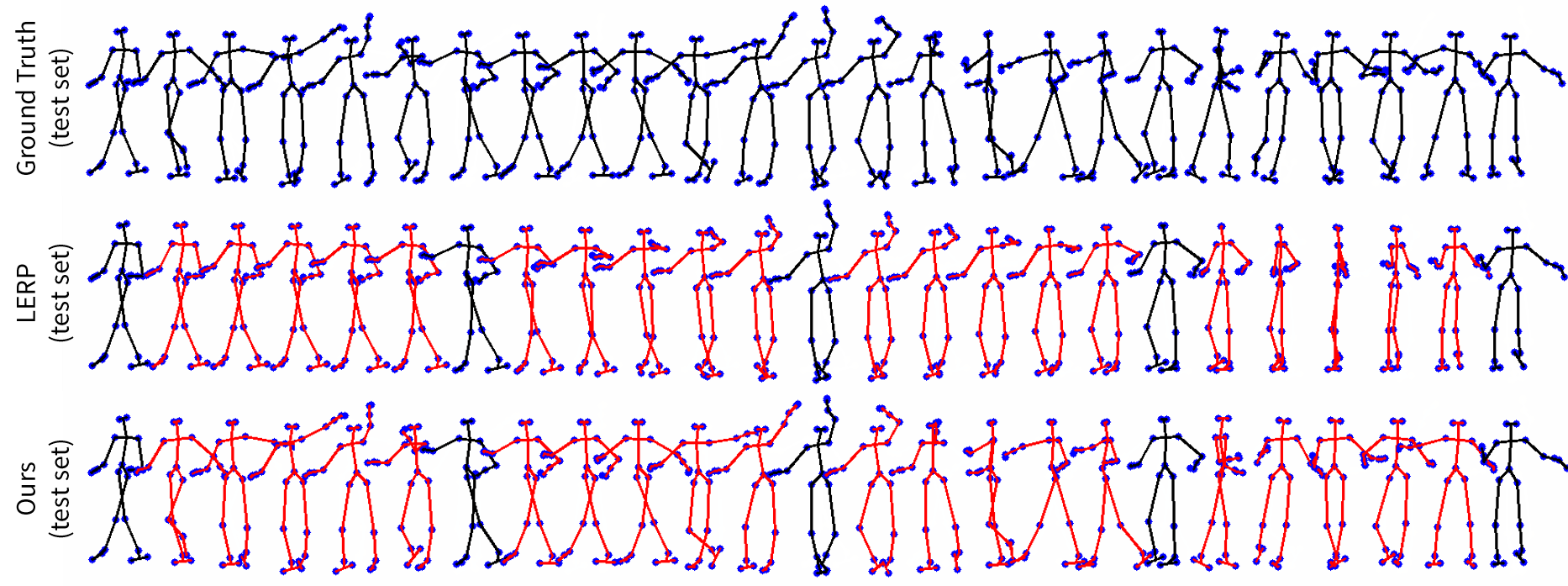}
  \caption{Our transformer-based infilling results and linear interpolation based results on the anidance test set. The fist row is the ground truth. In the rest rows, red skeletons are predicted and black ones are input keyframes.}
  \label{fig:infilling_test}
\end{figure*}

\begin{figure*}[ht]
  \centering
  \includegraphics[width=0.9\linewidth]{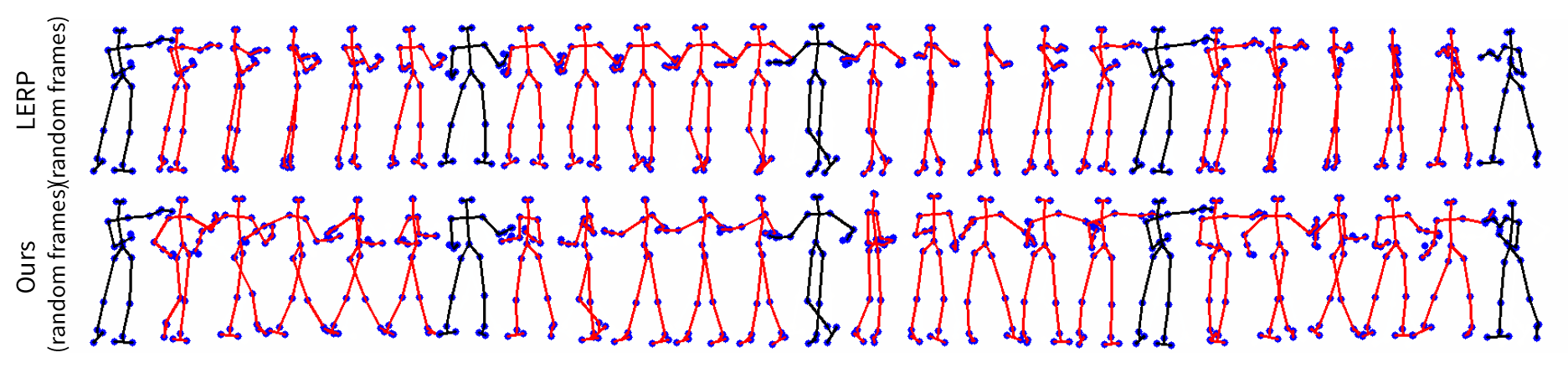}
  \caption{Our transformer-based infilling results and linear interpolation based results on the anidance test set (In this experiment, keyframes are randomly chosen from the test set with a random order for simulating in-the-wild scenario).}
  \label{fig:infilling_wild}
\end{figure*}

\subsection{Motion In-betweening} \label{Sec:in-betweening}

We evaluate our method on the LaFAN1 dataset~\cite{harvey2020robust} for the motion in-betweening task. This dataset contains 496,672 frames performed by 5 motion subjects and are recorded by using Mocap in 30Hz. The training set and the test set are clearly separated, where the test set contains motions from subject No.5 only. Since the originally captured motions are in very long sequences, motion windows are introduced on this dataset, where the width of the window is set to 50 (65) frames, and the offset is set to 20 (25) frames for training (test). Finally, there are 20,212 and 2,232 windows for training and test, respectively. We train our model on LaFAN1 for 1,000 epochs with random initialization. We set the maximum learning rate to $10^{-3}$, and set the weight decay to 0.75 every 200 epochs. We further adopt a warm-up strategy for the first 50 training epochs, where the learning rate goes up from 0 to the maximum learning rate gradually. Since our method can take in arbitrary inputs, we set the transition length of in-betweening to 5$\sim$39 (since the maximum length is 39). For the unknown frames, before feeding them to our model, we interpolate them based on the nearest two keyframes by linear interpolation (LERP) and spherical linear interpolation (SLERP).

When the training stops, we evaluate our method on the test set of LaFAN1. Tab \ref{tab:inbetweening} shows the evaluation result, where interpolation and zero-velocity are used as our naive baselines~\cite{harvey2020robust}. During the evaluation, only the first 10 keyframes and another keyframe at the frame $10+L+1$ are given, where $L$ is a predefined transition length. We keep the same configuration with Harvery \etal, where transition lengths are set to 5, 15, and 30. We evaluate our method under both local and global coordinate systems.

\begin{table}[ht]
\centering
\caption{Infilling results on anidance dataset~\cite{tang2018anidance} (A lower score indicates a better performance).}
\begin{tabularx}{0.8\linewidth}{l@{\extracolsep{\fill}} ccc}
&\multicolumn{3}{c}{\textbf{L2P}} \\
\cmidrule(lr){2-4}
Length & 5 & 15 & 30 \\
\cmidrule(lr){1-4}
Zero-Vel & 2.34 & 5.12 & 6.73\\
Interp & 0.94 & 3.24 & 4.68 \\
Ours (full)  & \bf{0.84} & \bf{1.46} & \bf{1.64} \\
\bottomrule
\end{tabularx}%
\label{tab:infilling}
\end{table}%

\begin{table*}[ht]
\centering
\caption{Blending results of our new dance dataset (A lower score indicates a better performance).}
\begin{tabularx}{\textwidth}{l@{\extracolsep{\fill}} ccccccccc}
&\multicolumn{3}{c}{\textbf{L2Q}} &\multicolumn{3}{c}{\textbf{L2P}} &\multicolumn{3}{c}{\textbf{NPSS}} \\
\cmidrule(lr){2-4}\cmidrule(lr){5-7}\cmidrule(lr){8-10}
Length & 8 & 16 & 32 & 8 & 16 & 32 & 8 & 16 & 32\\
\cmidrule(lr){1-10}
Zero-Vel & 2.17 & 2.67 & 3.22 & 3.68 & 5.15 & 7.52 & 0.2061 & 0.6004 & 1.7998 \\
Interp & 2.00 & 2.55 & 3.14 & 1.84 & 2.87 & 4.19 & 0.1948 & 0.5781 & 1.7218 \\
\cmidrule(lr){1-10}
Ours (global-native-Tpose)  & 1.62 & 2.03 & 2.48 & \emph{1.55} & \emph{2.32} & \emph{3.29} & 0.1906 & 0.5438 & 1.4758\\
Ours (global-standard-Tpose)  & 1.62 & 2.03 & 2.48 & 1.71 & 2.46 & 3.45 & 0.1906 & 0.5438 & 1.4758 \\
\bottomrule
\end{tabularx}%
\label{tab:blending}
\end{table*}%

We show in Tab \ref{tab:inbetweening} that our method achieves a high accuracy on the LaFAN1 dataset even when the transition length = 5. We can also see that there is a noticeable accuracy improvement when switching the generation mode from local to global. This may be because of the accumulative errors of rotations in local coordinates, and the previous method proposes to use the FK loss to reduce this error~\cite{harvey2020robust}. Our results suggest that the motion completion can be better solved in the global coordinate system, although the T-pose predicted in the global may not be accurate enough (can be further improved by IK loss). We further evaluate the Mixture Embedding (ME) and IK loss used in our method. We can see both the two strategies significantly improves the accuracy in all evaluation settings (L=5, 15, and 30).

Besides, Tab \ref{tab:speed} indicates that our method can achieve a very high inference speed on CPU device and can even run in real-time ($<$ 0.033s), which benefits from the non-autoregressive design.

\subsection{Dance in-filling}

Next, we evaluate our method on the Anidance dataset~\cite{tang2018anidance}. In this dataset, four types of dance movements (Cha-cha, Tango, Rumba, and Waltz) are captured in the global coordinate system. This dataset was originally designed for music-to-dance generation and contains 61 independent dance fragments with 101,390 frames.

We apply similar evaluation settings on this dataset, where the time window is set to 128 frames and the offset is set to 64. 20\% dance fragments are randomly selected as the test set, and thus there are 1,117 sequences in the training set and 323 sequences in the test set. We train our model on this dataset with 3000 epochs, and the rest of the configurations are kept the same with Sec.~\ref{Sec:in-betweening}. We set the interval of keyframes to 5$\sim$30 for in-filling, which means that there are only 5 keyframes are given to the model when their interval is 30, and the initial transition between keyframes are also interpolated by LERP. 

Tab \ref{tab:infilling} shows the evaluation results of our method on the test set of Anidance. The transition lengths are set to 5, 15, and 30 respectively. We evaluate our method under the global coordinate system since the anidance dataset contains global positions only. Similar to the results in the in-betweening task, our method can achieve high completion accuracy on long-term completion, and can also outperform LERP in short-term completion. Our method can also handle in-the-wild input (random keyframes) very well as shown in Fig.\ref{fig:infilling_test} and \ref{fig:infilling_wild}, where our method complete much more meaningful dance movements than the LERP. 

\begin{figure}
  \centering
  \includegraphics[width=0.9\linewidth]{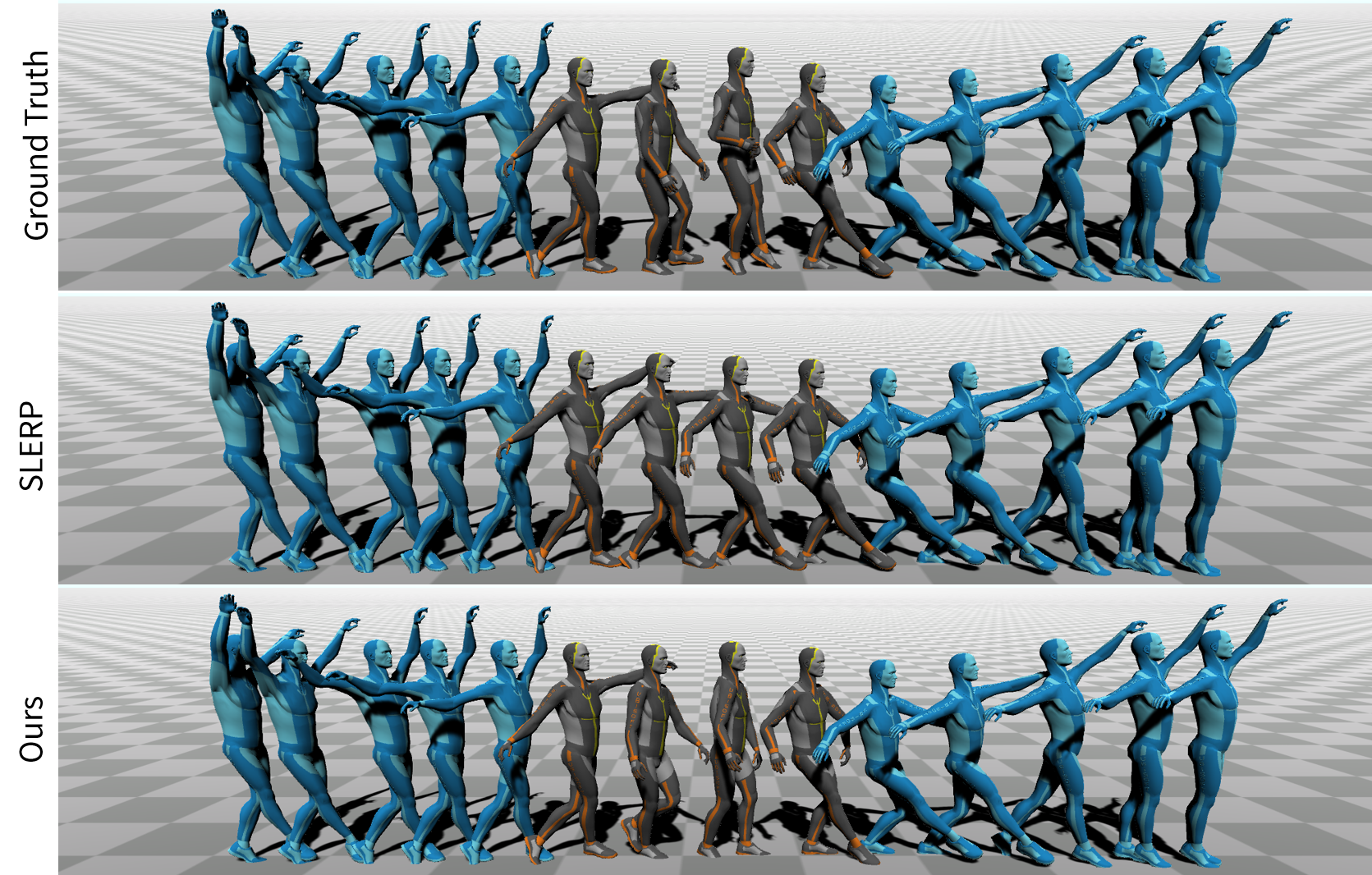}
  \caption{Results of different blending methods on our dataset with window width = 16. Our results are much closer to the ground truth.}
  \label{fig:blending}
\end{figure}

\subsection{Dance blending}

The above experiments show that our method can apply very well to the current public datasets. For further evaluating our method, we test our method on a very challenging dataset we built. Our dataset contains more complex and diverse dance movements. In our dataset, the time window is set to 64 and the offset is set to 32, and there are finally 3,220 sequences in the training set and 400 sequences in the test set. We keep the same configuration with Sec.~\ref{Sec:in-betweening}, but use pre-trained weights from Sec.~\ref{Sec:in-betweening} for initializing.

We set the blending window to 5$\sim$32 frames. For example, when the window is 32, the first and last 16 frames are given as keyframes and rests are masked. Tab \ref{tab:blending} shows our evaluation results with the window width = 8, 16, 32. Fig. \ref{fig:blending} shows our results with the window width = 16, which is a commonly-used setting ($\sim 0.5$s) in auto-blending. We can see our method can still achieve very high quality results on this challenging dataset.

\subsection{Discussion}

Our method can work on both global and local coordinate systems. When directly predicting poses in the global coordinate system, the bone length may not be well-controlled, especially in those games with relative coordinate systems (e.g., local root and rotation), although it may bring a higher accuracy. For example, in the last two rows of Tab \ref{tab:blending}, we evaluate our result by using standard T-pose criteria and the penultimate one (global-native-Tpose) respectively. The global-native-Tpose is slightly more close to the ground truth but is hard to directly apply to local coordinate applications. We also notice that the global rotation may have discontinuity when the dance is very fast (caused by SLERP). However, this is not a serious problem on the LaFAN1 dataset since the motions in this dataset are slow. We will also further investigate this interesting problem.

\section{Conclusion}

In this paper, we propose a simple but effective method to solve motion completion problems under a unified framework. In our method, a standard transformer encoder is introduced to handle the arbitrary sequence input, and a mixture embedding is introduced to better encode temporal information of multiple input types. Our method can predict multiple missing frames in a single forward propagation at inference time rather than running in an auto-regressive manner. Experimental results show that our method can be well applied to different motion completion modes, including in-betweening, in-filling and blending, and achieves a new state of the art accuracy on the LaFAN1 dataset.

{\small
\bibliographystyle{ieee_fullname}
\bibliography{ssmct}
}

\end{document}